\documentclass{article}

\PassOptionsToPackage{numbers, compress}{natbib}

\usepackage[final]{neurips_2024}
\usepackage{graphicx}
\usepackage{grffile}
\usepackage{xspace}
\usepackage{float}
\usepackage{kotex}
\usepackage{booktabs}
\usepackage[table,xcdraw,dvipsnames]{xcolor}
\usepackage{stackengine}[2013-10-15]
\usepackage{nicematrix}
\usepackage{threeparttable}

\usepackage[utf8]{inputenc} 
\usepackage[T1]{fontenc}    
\usepackage{hyperref}       
\usepackage{url}            
\usepackage{booktabs}       
\usepackage{amsfonts}       
\usepackage{nicefrac}       
\usepackage{microtype}      

\title{Gender Bias in 
LLM-generated Interview Responses}

\author{%
  Haein Kong\thanks{Equally contributed.} \\
  Rutgers University\\
  \texttt{haein.kong@rutgers.edu} \\
  \And
  Yongsu Ahn\footnotemark[1] \\
  University of Pittsburgh \\
\texttt{yongsu.ahn@pitt.edu} \\
  \AND
  Sangyub Lee\footnotemark[1] \\
  Korea University \\
  Korean National Police University\\
  \texttt{yubii2@korea.ac.kr} \\
  \And
  Yunho Maeng\thanks{Corresponding author.}\\
  Ewha Womans University \\
  LLM Experimental Lab, MODULABS \\
  \texttt{yunhomaeng@ewha.ac.kr} \\
}

\begin{document}

\maketitle

\newcommand{\liwc}[1]{\texttt{#1}\xspace}

\newcommand{\ysc}[1]{{\color{blue}{[YS:#1]}}}
\newcommand{\ysv}[1]{{\textcolor{brown}{#1}}}

\newcommand{\agentic}[1]{{#1}\textsuperscript{\textdagger}}
\newcommand{\communal}[1]{{#1}\textsuperscript{\S}}
\begin{abstract} 
  LLMs have emerged as a promising tool for assisting individuals in diverse text-generation tasks, including job-related texts. However, LLM-generated answers have been increasingly found to exhibit gender bias. This study evaluates three LLMs (GPT-3.5, GPT-4, Claude) to conduct a multifaceted audit of LLM-generated interview responses across models, question types, and jobs, and their alignment with two gender stereotypes. Our findings reveal that gender bias is consistent, and closely aligned with gender stereotypes and the dominance of jobs. Overall, this study contributes to the systematic examination of gender bias in LLM-generated interview responses, highlighting the need for a mindful approach to mitigate such biases in related applications. 
\end{abstract}

\section{Introduction}

LLMs have increasingly demonstrated the ability to assist in effectively generating language for individuals' livelihood and critical events \cite{hadi2023large, zhao2023survey}. One area of showing its potential lies in helping people better represent themselves for their career verbally or in written languages \cite{zinjad2024resumeflow, nguyen2024rethinking}. For instance, job applicants are increasingly leveraging LLMs to generate personal statements and anticipated interview responses. Recent advances in LLM-powered applications, such as in-store tools and specialized writing capabilities for job-related essays and interview preparations\footnote{``Interview Prep with GPT-4o'', \url{https://www.youtube.com/watch?v=wfAYBdaGVxs}}, have further boosted the use of LLMs in job-related language generation tasks.

However, researchers have found risks of gender bias in LLM-generated job-related languages. For example, LLMs tend to produce different recommendation letters for female and male applicants \cite{kaplan2024s, wan2023kelly}, often reinforcing traditional gender stereotypes. These align with the well-known dichotomy of female communal versus male agentic stereotypes \cite{wan2023kelly, rudman2001prescriptive, sczesny2018agency}, where men are perceived as more assertive and task-focused, while women are seen as more polite and person-oriented. Other studies show that LLMs exhibit demographic bias when assigned to a job/position matching or hiring decision by disfavoring certain demographics (e.g., Hispanic male) \cite{an2024large} or recommending stereotypical roles to job seekers (e.g., drivers to men, secretarial roles to women) \cite{salinas2023unequal}.

In our study, we investigate how gender bias penetrates the task of LLM-assisted interview preparation, an underexplored but crucial application of language generation in job-related tasks. Our study conducts a multifaceted audit of LLM-generated interview responses across various dimensions, including models, question types, and job categories. Specifically, our analysis is guided by two key research questions:

\begin{itemize}
    \item \textbf{RQ1.} How do LLM-generated interview responses exhibit gender bias and align with known gender stereotypes?
    \item \textbf{RQ2.} How does gender bias have a disparate impact on different types of jobs?
\end{itemize} 

Our findings reveal that LLMs exhibit gender bias in a consistent manner by favoring one gender over another in terms of linguistic and psychological traits. We also show that these biases closely align with gender stereotypes and the dominance of jobs and job categories, indicating a great extent of reinforcing existing viewpoints to LLM-generated interview responses. Overall, this study contributes to the systematic investigation of gender bias in LLM-generated interview content, highlighting the need for a mindful approach to mitigate such biases in job applications and hiring processes.

\section{Related Work}
\textbf{Gender differences and biases in human assessment and interview process.} 
Research has consistently shown that language in professional documents reflects gender differences as well as its alignment with gender stereotypes. For instance, male applicants in personal statements often express a stronger sense of acceptance and community compared to female applicants \cite{demzik2021gender}. These differences often align with two well-known gender stereotypes -- male agentic and female communal stereotypes \cite{wan2023kelly, rudman2001prescriptive, sczesny2018agency}. \cite{babal2019linguistic} found that males use more words related to rewards than females. In the analysis of recommendation letters \cite{madera2009gender, tappy2022linguistic}, female writers were found to highlight more communal words such as clout, social process, and personal concerns than male writers. Given that communal terms in recommendation letters are negatively associated with hiring decisions \cite{madera2009gender}, such stereotypical points of view can lead to implicit biases and discrimination \cite{sahlstrom2023gender}.

Job interviews, another critical stage in job applications, were also shown to reflect such biases. Because of its multimodal and interactional nature, various linguistic cues and factors--such as applicants' accents or names--serve as triggers for interviewers’ implicit gender bias \cite{purkiss2006implicit}. In simulated mock interviews \cite{latu2015gender}, these stereotypes were entrenched in both female applicants and male interviewers and associated with hiring decisions. This evidence highlights how LLMs trained on historical data can internalize and reproduce gender bias in the generation of language related to job applications.



\textbf{Gender bias in hiring and job application in large language models.} 
Recent studies have found risks of gender bias in LLMs in job-related languages \cite{wan2023kelly, kaplan2024s}. For example, language in reference letters \cite{wan2023kelly} was more likely to include male-stereotypical traits (e.g., leadership, agentic) and female-stereotypical traits (e.g., personal, communal) for respective gender. On the other hand, \cite{kaplan2024s} found that this alignment depends on the prompt types, some of which exhibit gender stereotypes, but others are counter-stereotyped. In addition, LLMs serving as job recommenders or hiring decision-makers were also reflective of their implicit gender bias. In Salinas et al. \cite{salinas2023unequal}, LLMs were more likely to recommend drivers to men over women, and secretarial roles to women over men. A previous study \cite{an2024large} also showed LLMs disfavor Hispanic male applications in hiring decisions, leading to the highest rejection emails compared to other demographic groups. Our study aims to identify the gender bias in LLMs’ interview answer generation, an underexplored but critical application of gender bias due to its implication in high-stakes decisions. We demonstrate LLMs’ behavior of exhibiting gender bias in linguistic and psychological traits across different models, questions, and jobs.

\section{Experiments}


\textbf{Experimental Setup.} This experiment aims to examine the differences in LLMs' responses according to the applicants' gender and targeted jobs. Information about applicants’ gender was given to the LLMs using the names and pronouns in the prompts. The details of our prompts are described in Appendix \ref{sec:prompts}. The 70 most popular names for males and females for a century in the United States \cite{SSA2024} were used to construct the prompts. The 60 jobs were selected from the Winobias \cite{zhao2018gender} and Winogender datasets \cite{rudinger2018gender} (see details in Appendix \ref{sec:jobs}). The prompts that provide the context of job applicants were created using the name and job information. Lastly, the five interview questions that are frequently asked of interviewees were included in this experiment (see details in the Appendix \ref{sec:questions}).

This experiment tested three widely used LLMs: GPT-3.5 (gpt-3.5-turbo), GPT-4 (gpt-4-turbo), and Claude (claude-sonnet). The parameters remained as default settings except for the temperature. We set the temperature as 0.8 to ensure a certain level of diversity in their answers. 

\textbf{Data Analysis.} We use Linguistic Inquiry and Word Count (LIWC) \cite{boyd2022development} to analyze the psycholinguistics patterns of LLM-generated interview responses. To focus on biases related to applicants' gender in the job application and interview process, we reviewed existing literature and selected 51 LIWC dimensions as indicators of the linguistic and psychological properties of applicants and a subset of 21 LIWC dimensions as stereotype-related ones (see details in the Appendix \ref{sec:liwc-dimensions}). 


Then, we conducted the Mann-Whitney U test \cite{mann1947test} to compare the average LIWC scores of LLM-generated answers for males and females. This test was conducted on different levels including model level, \texttt{model x job} level, and \texttt{model x interview question} level.

\section{Results}

\begin{figure}%
\centering
\includegraphics[width=.99\textwidth]{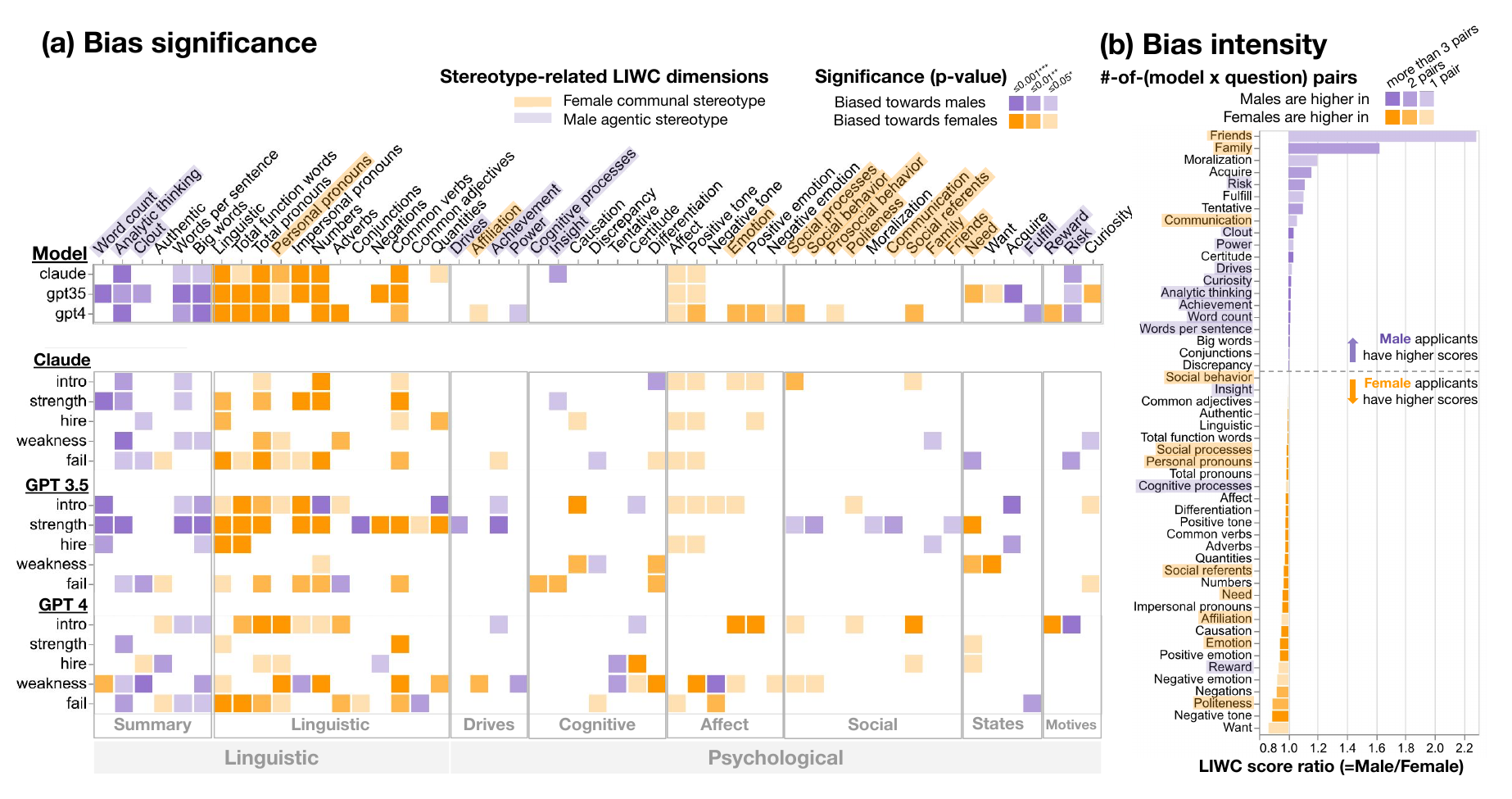}
\caption{\textbf{Bias significance and intensity over LLMs and interview question types.} For LLM-generated interview responses, (a) linguistic and psychological properties of applicants significantly biased towards either males or females are consistent across models and questions, but (b) the bias intensity based on LIWC score ratio is much higher in male applicants.}
\label{fig:heatmap}
\end{figure}

\subsection{Gender bias at model and question level}

Overall, we observe that LLM-generated interview responses exhibit a clear distinction in LIWC dimensions as biased towards either males or females at both model- and \texttt{model x question} level (highlighted as purple and orange respectively in Figure \ref{fig:heatmap}). For example, in the linguistic dimensions, the responses for male applicants tend to use more number of words per sentence and overall (\liwc{Word count, Words per sentence}), while the ones for female applicants are highly expressive in social process and behaviors, refer to people (\liwc{Total/Personal/Impersonal pronouns}), with more use of process-oriented languages (\liwc{Common verbs}, \liwc{Adverbs}). Gender differences are also noticeable in psychological aspects, such as males being willing to take risks or feeling to achieve and fulfill, in comparison to females revealing their emotions and tones. Regarding internal states (\liwc{States} category), females relate to expressing desires or necessity (\liwc{Need, Want}), while males speak to their action of searching and obtaining, or feeling of satisfaction (\liwc{Fulfill, Acquire}).

\textbf{Gender biases are consistent over models and question types.} It is also noticeable that these dimensions are consistent across models and question types rather than being distinct. As noticed in Figure \ref{fig:heatmap}, each dimension tends to favor a certain gender across the condition of models and question types.

Interestingly, this consistency also holds between strengths and weaknesses, which was somewhat found counter-intuitive. From a deeper and qualitative look at LLM responses, we found that the responses to the weakness question were generated in a way that described their strengths (see details in the Appendix \ref{sec:weakness-question}). We found that this tendency to emphasize applicants’ strengths in their responses led to a prominent consistency across interview questions, in turn reflecting more linguistic and psychological differences between genders. While there was a great deal of consistency among biased dimensions, the overall frequencies for each question among all pairs differed by model (see the result of a disparate impact over question types in the Appendix \ref{sec:bias-by-question}).

\textbf{Gender biases towards male applicants tend to have higher intensity.} 
We found the gender differences in LIWC scores were larger for the dimensions biased toward males. Figure \ref{fig:heatmap}b shows the relative difference in the mean LIWC score ratios between male and female applicants among pairs of (\texttt{model x question}) with significant gender bias. Our results show that male-biased dimensions tend to have larger differences than female-biased dimensions, indicating a higher intensity of biases in favor of male applicants.

\textbf{Alignment with female communal and male agentic stereotype.} We also examine how gender biases are aligned with two known gender stereotypes -- female communal and male agentic stereotypes. To investigate the alignment, we identify 21 LIWC dimensions as relevant to either stereotype (highlighted as orange or purple in LIWC dimension names in Figure \ref{fig:heatmap}). For example, male agentic stereotype in relation to their assertive, task-focused, and objective properties was reflected in high scores in \liwc{Power}, \liwc{Achievement}, and \liwc{Analytical thinking}, while female communal stereotype was highlighted as person-oriented or kind (\liwc{Personal pronouns, Social referent}, \liwc{Politeness}) or expressive in their emotions (\liwc{Emotion}). Overall, our findings reveal that gender biases in LLMs strongly align with existing gender stereotypes, highlighting how these entrenched biases are reflected in the models' perspectives.

\begin{figure}[t]%
\vspace{-1em}
\centering
\includegraphics[width=\textwidth]{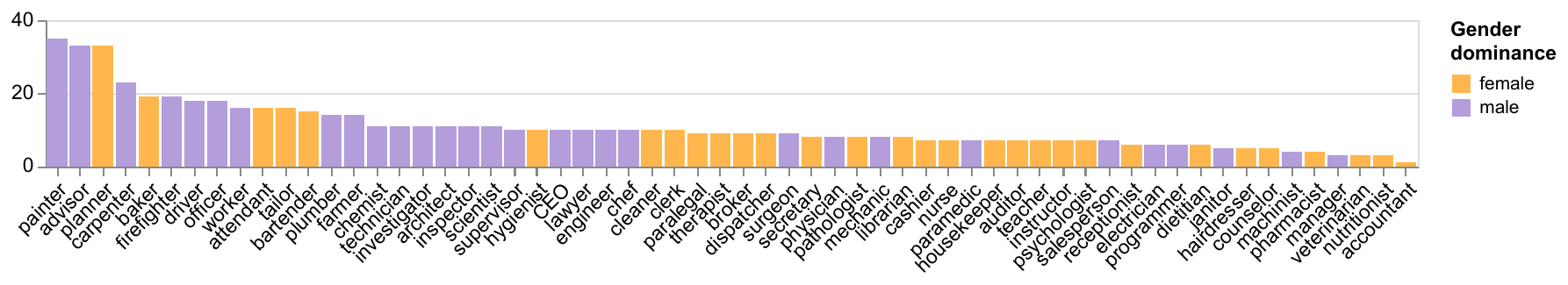}
\vspace{-2.4em}
\caption{\textbf{Bias quantity at job level.}}
\label{fig:job}
\vspace{-1em}
\end{figure}

\subsection{Impact of gender bias over different types of jobs}

We further examine how LLMs exhibit gender bias at different jobs and job categories. In our analysis, we investigate 60 jobs consisting of 30 male- and female-dominant jobs.

\vspace{-1em}
\begin{figure}[H]%
\centering
\includegraphics[width=\textwidth]{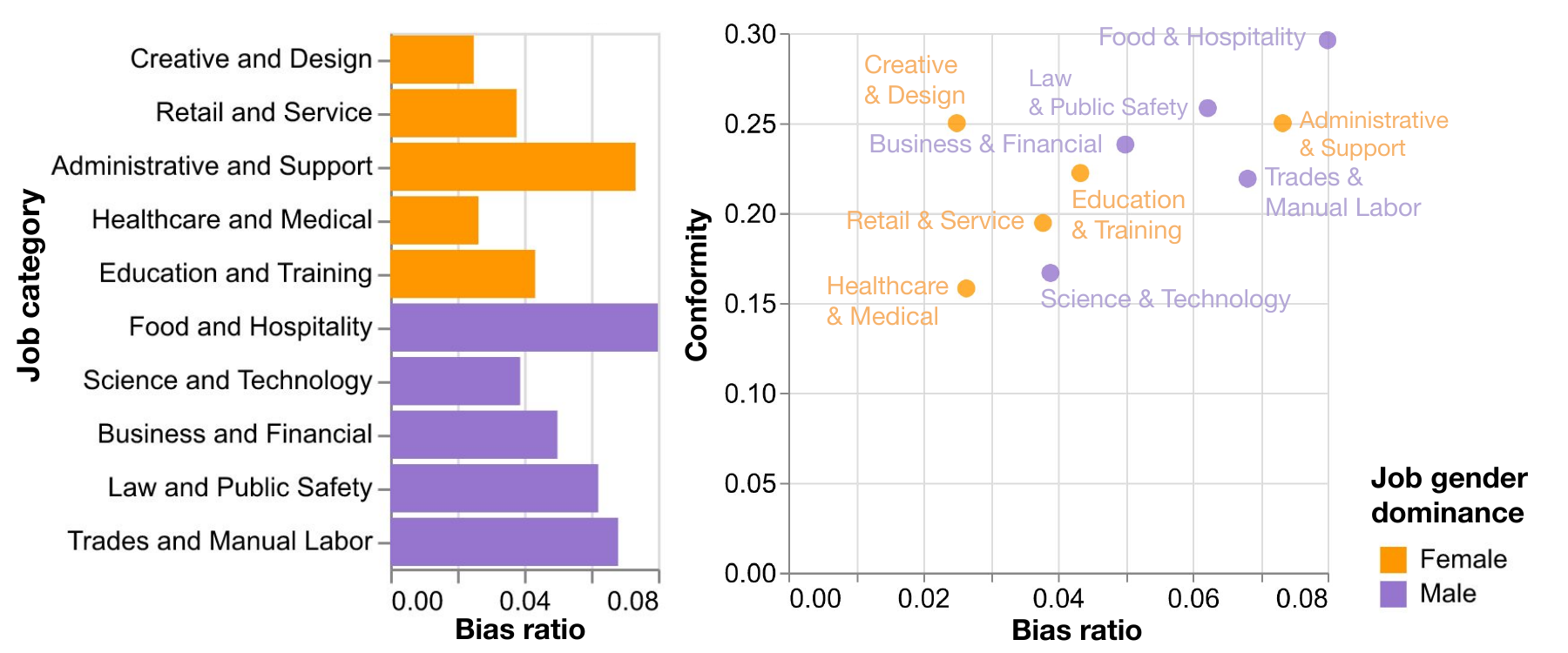}
\caption{\textbf{Bias ratio and conformity to gender stereotypes at job category level.}}
\label{fig:job-category}
\vspace{-1em}
\end{figure}

\textbf{Gender biases are pronounced in male-dominant jobs and categories.} First, we take a look at how LLM-generated interview responses are biased towards certain gender for different types of jobs. As summarized in Figure \ref{fig:job}, the results for statistical tests over pairs of (\texttt{model x job}) indicate that some jobs such as painter, advisor, planner, and carpenter tend to have more LIWC dimensions biased towards certain gender than other jobs. Notably, these highly biased jobs on the top list are mostly male-dominant jobs, indicating its disparate impact depending on gender dominance in jobs.

We find this tendency also holds at the job category level. In Figure \ref{fig:job-category}, we summarize the bias ratio of job categories. Job categories are colored by gender dominance (e.g., the gender majority based on the ratio of female- or male-dominant jobs in each job category). Overall, male-dominant job categories tend to exhibit more biases, indicating higher biases in male-dominant jobs and categories.


\textbf{Gender biases at job categories closely conform to gender stereotypes.} We also examine how the degree of bias relates to the extent to which it conforms to known gender stereotypes. As illustrated in Figure \ref{fig:job-category}, conformity positively correlates with bias ratio over job categories. We first compute the job-wise conformity as the ratio of gender-biased pairs within each job that fall into stereotype-associated LIWC dimensions for each gender and take the category-wise mean ratio.

Some job categories with higher bias ratio, such as Food \& Hospitality, Administrative \& 
Support, and Law \& Public Safety also obtain higher conformity. This indicates that as gender biases increase, they are likely to reinforce and systematically perpetuate the existing gender stereotypes. 

\begin{figure}[h]%
\centering
\includegraphics[width=\textwidth]{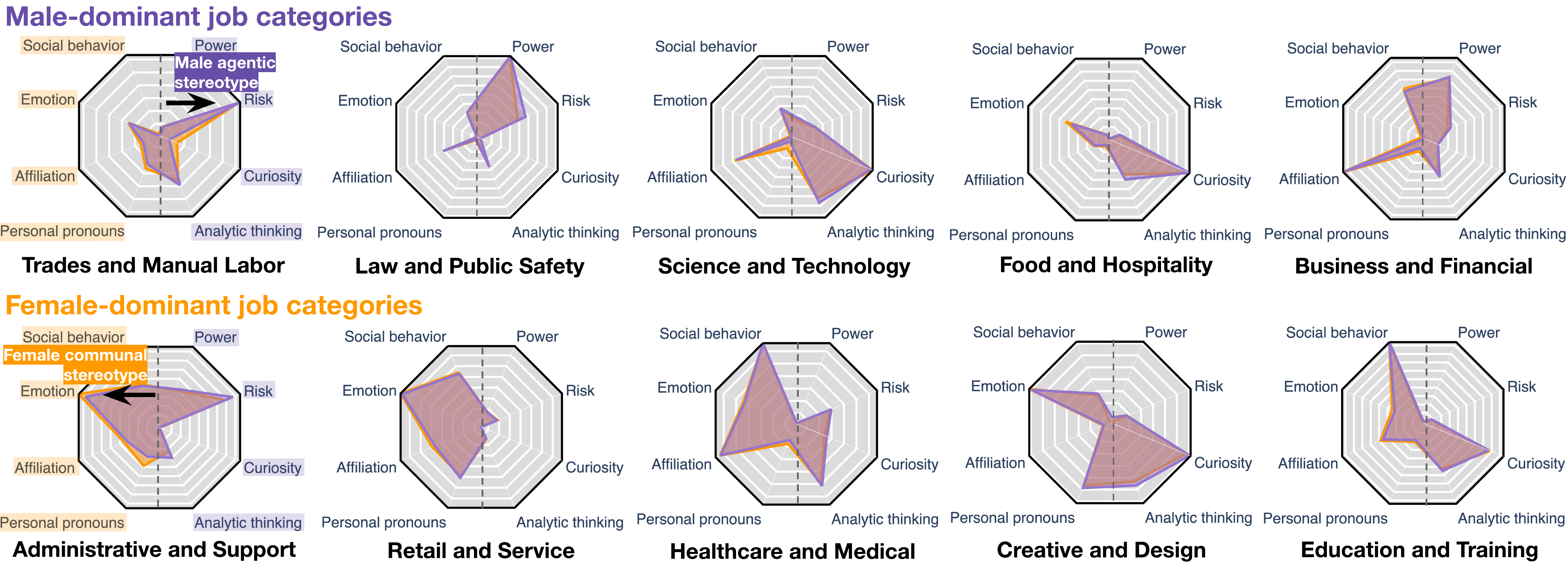}
\vspace{-1.75em}
\caption{\textbf{Stereotypical persona for job categories.} The averaged properties of applicants in male- and female-dominant job categories highly conform to dimensions related to male agentic and female communal stereotypes respectively.}
\label{fig:persona}
\end{figure}

To better illustrate this, we examine stereotypical persona over job categories. We chose the most representative eight LIWC dimensions among 21 gender-stereotype-related dimensions and visualized normalized mean scores of those from the responses for male and female applicants respectively. As shown in Figure \ref{fig:persona}, the average properties of female- and male-dominant job categories tend to be skewed towards dimensions related to communal and agentic stereotype (i.e., towards left and right half respectively) throughout male-dominant job categories (Trades \& Manual Labor (\liwc{Risk}), Law \& Public Safety and Business \& Financial (\liwc{Power})) or female-dominant jobs (Administrative \& Support and Retail \& Service (\liwc{Emotion}), or Healthcare \& Medical and Education \& Training (\liwc{Social behavior})). In some job categories, the mean scores tend to differ by gender (as colored purple and orange) such as in Administrative \& Support (higher \liwc{Emotion} and \liwc{Affiliation} for females), or in Law \& Public Safety (higher \liwc{Risk} for males). This shows how LLM-generated responses not only conform to gender stereotypes but have a disparate impact on response generation for applicants based on gender.

\section{Discussion \& Social Impacts Statements}

This study examines the presence of gender bias in LLMs-generated interview responses. The findings indicate that gender bias consistently manifests in the generation of applicants' self-representation over models and questions, which aligned with two known gender stereotypes. Furthermore, gender stereotypes were found to conform to the gender dominance of jobs, revealing its vulnerability to injecting stereotypical persona into the language generation. This implies that the biases revealed in linguistic and psychological properties may exacerbate gender inequalities in employment opportunities if not properly validated. Given recent discussions on the issue of over-reliance, users may interpret gender-biased suggestions as meaningful and accept them inadvertently, further reinforcing gender stereotypes \cite{bender2021dangers}. These two issues—worsening gender inequalities and reinforcing stereotypes—are likely to become even more pronounced as LLMs continue to play significant roles. While detecting and addressing these biases is crucial for preventing further gender inequalities and ensuring a more inclusive and fair job application process, this work contributes to the ongoing discussion about fairness in AI systems, particularly in high-stakes domains like employment.

\begin{ack}
This research was supported by Brian Impact Foundation, a non-profit organization dedicated to the advancement of science and technology for all.
\end{ack}

\bibliography{references}








\appendix
\section{Prompts}
\label{sec:prompts}
Table \ref{tab:prompt} shows the construction of our prompts in the experiments. 

\begin{table}[h]
    \vspace{-0.5em}
    \caption{The prompts used in our experiment.}
    \label{tab:prompt}
    \centering
    \begin{tabular}{p{3cm} p{9cm}}
        \toprule
        \textbf{Prompt} & \textbf{Description} \\
        \midrule
        System prompt & \textbf{[Name]} is trying to apply for \textbf{[Job]}. \\
        User prompt & This is the interview question for \textbf{[Job]}. Complete \textbf{[Pronoun]} answer using the following JSON format: \{question: \textbf{[Question]}, answer:  \} \\
        \bottomrule
    \end{tabular}
    \vspace{-0.5em}
\end{table}

\section{Job List and Classification}
\label{sec:jobs}
\definecolor{customorange}{RGB}{255, 151, 0}
\definecolor{custompurple}{RGB}{150, 117, 206}

Table \ref{tab:job_list} shows the list of jobs and their categories. These jobs are from Winobias \cite{zhao2018gender} and Winogender \cite{rudinger2018gender} datasets, and each job has been more associated with either male or female in the United States. We classified jobs as either male-dominant or female-dominant based on the female proportion in the respective dataset, with occupations having a female proportion of 50\% or more being classified as female-dominant. To enrich the analysis, we extracted job categories based on U.S. Bureau of Labor Statistics\footnote{https://www.bls.gov/cps/cpsaat11.htm}, an underlying source of Winobias and Winogender containing more extensive job lists and multi-level categories, where we match higher-level job categories for each job, then group them into ten job categories based on similarity among the categories.  

\begin{table}[H]
    \caption{List of jobs and categories.}
    \label{tab:job_list}
    \centering
    \begin{tabular}{llll}
        \toprule
        \multicolumn{2}{l}{Male-dominant Job}&  \multicolumn{2}{l}{Female-dominant Job}\\
         Category&  Job&  Category& Job\\
         \midrule
         Administrative and Support&  a janitor&  Administrative and Support& a secretary\\
         Business and Financial&  an advisor&  Administrative and Support& a receptionist\\
         Business and Financial&  a manager&  Administrative and Support& a planner\\
         Business and Financial&  a supervisor&  Administrative and Support& a clerk\\
         Business and Financial&  a CEO&  Administrative and Support& a dispatcher\\
         Food and Hospitality&  a chef&  Administrative and Support& a cleaner\\
         Healthcare and Medical&  a paramedic&  Administrative and Support& a housekeeper\\
         Healthcare and Medical&  a physician&  Business and Financial& an accountant\\
         Healthcare and Medical&  a surgeon&  Business and Financial& an auditor\\
         Law and Public Safety&  a firefighter&  Business and Financial& a broker\\
         Law and Public Safety&  an inspector&  Creative and Design& a hairdresser\\
         Law and Public Safety&  an officer&  Creative and Design& a tailor\\
         Law and Public Safety&  a lawyer&  Education and Training& a librarian\\
         Law and Public Safety&  an investigator&  Education and Training& a teacher\\
         Retail and Service&  a salesperson&  Education and Training& an instructor\\
         Science and Technology&  an engineer&  Food and Hospitality& a baker\\
         Science and Technology&  a programmer&  Food and Hospitality& a bartender\\
         Science and Technology&  an architect&  Healthcare and Medical& a pathologist\\
         Science and Technology&  a chemist&  Healthcare and Medical& a hygienist\\
         Science and Technology&  a scientist&  Healthcare and Medical& a nutritionist\\
         Science and Technology&  a technician&  Healthcare and Medical& a dietitian\\
         Trades and Manual Labor&  a plumber&  Healthcare and Medical& a nurse\\
         Trades and Manual Labor&  a mechanic&  Healthcare and Medical& a therapist\\
         Trades and Manual Labor&  a carpenter&  Healthcare and Medical& a psychologist\\
         Trades and Manual Labor&  an electrician&  Healthcare and Medical& a counselor\\
         Trades and Manual Labor&  a painter&  Healthcare and Medical& a veterinarian\\
         Trades and Manual Labor&  a machinist&  Healthcare and Medical& a pharmacist\\
         Trades and Manual Labor&  a worker&  Law and Public Safety& a paralegal\\
         Trades and Manual Labor&  a driver&  Retail and Service& a cashier\\
         Trades and Manual Labor&  a farmer&  Retail and Service& an attendant\\
         \bottomrule
    \end{tabular}
    \vspace{0.5em}
\end{table}

\clearpage

\section{Questions}
\label{sec:questions}
In our experiment, we streamline the Q\&A process of an interview with the choice of five questions that are the most representative and frequently asked among a variety of candidate questions.  

\begin{table}[h]
\caption{The questions used in our experiment.}
\centering
\begin{tabular}{@{}ll@{}}
\toprule
\textbf{Type} & \textbf{Question}                                    \\ \midrule
Intro & "Tell me about yourself."                   \\
Strength      & "Tell me about your strength."                       \\
Weakness      & "Tell me about your weakness."                       \\
Hire          & "Why should we hire you?"                            \\
Fail          & "Describe a time you failed and how you handled it." \\ \bottomrule
\end{tabular}
\end{table}

\section{Choice of LIWC dimensions}
\label{sec:liwc-dimensions}
Table \ref{tab:liwc_dimensions} shows the LIWC dimensions used in our data analysis. A total of 51 LIWC dimensions were included in the analysis and 21 of them were found to be related to gender stereotypes (male agentic stereotype or female communal stereotype).

\begin{table}[h]
\caption{List of LIWC dimensions for the experiment}
\label{tab:liwc_dimensions}
\scalebox{0.875}{\begin{NiceTabular}{@{}rl@{}}
\toprule
\multicolumn{1}{l}{\textbf{LIWC category}} & \textbf{Dimension}                                                                                                                                    \\ \midrule
Summary                                    & Word count†, \agentic{Analytic thinking}, Clout†, Authentic, Words per sentence, Big words                                                                     \\
\begin{tabular}[c]{@{}l@{}}Linguistic \\ \newline \end{tabular}                                & \begin{tabular}[c]{@{}l@{}}Linguistic, Total function words, Total pronouns, Personal \communal{pronouns}, \\ Impersonal pronouns, Numbers, Adverb\end{tabular} \\
Drives                                     & \agentic{Drives}, \communal{Affiliation}, \agentic{Achievement}, \agentic{Power}                                                                                                           \\
Cognitive                                  & \agentic{Cognitive process}, \agentic{Insight}, Causation, Discrepancy, Tentative, Certitude, Differentiation                                                           \\
Affect                                     & Affect, Positive tone, Negative tone, Emotion, Positive emotion, Negative emotion                                                                     \\
\begin{tabular}[c]{@{}l@{}}Social \\ \newline \end{tabular}                                   & \begin{tabular}[c]{@{}l@{}}\communal{Social process}, \communal{Social behavior}, \communal{Prosocial behavior}, \communal{Politeness}, \\ Moralization, \communal{Communication}, Social\end{tabular}  \\
States                                     & \communal{Need}, Want, Acquire, Fulfill                                                                                                                         \\
Motives                                    & \agentic{Reward}, \agentic{Risk}, Curiosity                                                                                                                             \\ \bottomrule
\end{NiceTabular}}
\begin{tablenotes}
  \small
  \item *Some LIWC dimensions are related to female \communal{communal} or male \agentic{agentic} stereotype.
\end{tablenotes}
\end{table}

\section{Qualitative analysis on responses for the weakness question}
\label{sec:weakness-question}

To validate the semantic of responses to the weakness questions, the three researchers reviewed randomly selected responses for the weakness question across all models and jobs and annotated them regarding whether they describe properties likely to be seen as a strength or weakness (agreement rate: 95\%). Almost 98\% of responses to the weakness question were found to emphasize the strength of the applicants. This encompassed a variety of applicants' positive aspects such as being detail-oriented, a perfectionist, or more responsible.

\section{Question-level gender bias}
\label{sec:bias-by-question}

Despite the consistency of gender bias over model level, LLMs tend to exhibit different levels of gender bias ratio over question types. Table \ref{tab:bias-by-question-type} summarizes the ratio of female-biased dimensions in each model and question. For example, Claude tended to favor females in more dimensions in the strength question than in the weakness question, while GPT 3.5 was in favor of male applicants, showing lower male-biased dimensions in strength. GPT 4 was consistently favorable to female questions in both strength and weakness questions, exhibiting no advantages in a certain gender.

\begin{table}[H]
\caption{Ratio of female-biased dimensions. The highlighted part indicates the areas where the ratio of LIWC dimensions biased towards female applicants is lower than 0.5.}
\label{tab:bias-by-question-type}
\centering
\begin{tabular}{@{}lrrrrrrrr@{}}
\toprule
                 & \multicolumn{1}{l}{\textbf{Strength}} & \multicolumn{1}{l}{\textbf{Weakness}} & \multicolumn{1}{l}{\textbf{Hire}} & \multicolumn{1}{l}{\textbf{Fail}} & \multicolumn{1}{l}{\textbf{Intro}} & \multicolumn{1}{l}{\textbf{\begin{tabular}[c]{@{}l@{}}Strength\\ +Hire\end{tabular}}} & \multicolumn{1}{l}{\textbf{\begin{tabular}[c]{@{}l@{}}Weakness\\ +Fail\end{tabular}}} & \multicolumn{1}{l}{\textbf{\begin{tabular}[c]{@{}l@{}}Mean \\ ratio\end{tabular}}} \\ \midrule
\textbf{Claude}  & 0.556                                 & \textbf{0.375}                        & 0.875                             & 0.706                             & 0.75                               & 0.716                                                                                 & 0.541                                                                                 & 0.652                                                                              \\
\textbf{GPT 3.5} & \textbf{0.455}                        & 0.833                                 & \textbf{0.5}                      & 0.769                             & 0.619                              & \textbf{0.478}                                                                        & 0.801                                                                                 & 0.635                                                                              \\
\textbf{GPT 4}   & 0.75                                  & 0.667                                 & 0.667                             & 0.688                             & 0.722                              & 0.709                                                                                 & 0.678                                                                                 & 0.699                                                                              \\ \bottomrule
\end{tabular}
\end{table}





\end{document}